\documentclass[preprint,5p, twocolumn]{elsarticle}
\usepackage{listings}
\usepackage{amsmath}
\usepackage{algorithm}
\usepackage{algpseudocode}
\usepackage{tabularx}
\usepackage{caption}
\usepackage{subcaption}
\usepackage[dvipsnames]{xcolor}
\usepackage{multirow}
\usepackage{booktabs}

\definecolor{codegreen}{rgb}{0,0.6,0}
\definecolor{codegray}{rgb}{0.5,0.5,0.5}
\definecolor{codepurple}{rgb}{0.58,0,0.82}
\definecolor{backcolour}{rgb}{0.95,0.95,0.92}

\lstdefinestyle{mystyle}{
    backgroundcolor=\color{backcolour},   
    commentstyle=\color{codegreen},
    keywordstyle=\color{magenta},
    numberstyle=\tiny\color{codegray},
    stringstyle=\color{codepurple},
    basicstyle=\ttfamily\footnotesize,
    breakatwhitespace=false,         
    breaklines=true,                 
    captionpos=b,                    
    keepspaces=true,                 
    numbers=left,                    
    numbersep=5pt,                  
    showspaces=false,                
    showstringspaces=false,
    showtabs=false,                  
    tabsize=2
}
\newcommand{\ttheta}{{\boldsymbol{\theta}}}

\newcommand{\rlow}{{r_l}}
\newcommand{\rhigh}{{r_h}}

\newcommand{\figref}[2][]{{Figure~\ref{#2}#1}}
\newcommand{\tabref}[1]{{Table~\ref{#1}}}
\newcommand{\secref}[1]{{Section~\ref{#1}}}
\newcommand{\appref}[1]{{\ref{#1}}}

\definecolor{cream}{RGB}{222,217,201}
\definecolor{maroon}{cmyk}{0, 0.87, 0.68, 0.32}
\definecolor{halfgray}{gray}{0.55}
\definecolor{ipython_frame}{RGB}{207, 207, 207}
\definecolor{ipython_bg}{RGB}{247, 247, 247}
\definecolor{ipython_red}{RGB}{186, 33, 33}
\definecolor{ipython_green}{RGB}{0, 128, 0}
\definecolor{ipython_cyan}{RGB}{64, 128, 128}
\definecolor{ipython_purple}{RGB}{170, 34, 255}

\newcommand{\highlightone}[1]{\textcolor{NavyBlue}{#1}}
\newcommand{\highlighttwo}[1]{\textcolor{Orange}{#1}}
\newcommand{\highlightthree}[1]{\textcolor{ForestGreen}{#1}}
\newcommand{\highlightfour}[1]{\textcolor{BrickRed}{#1}}

\usepackage{amssymb}

\journal{Knowledge-Based Systems}

\begin{document}

\begin{frontmatter}



\title{SigOpt Mulch: An Intelligent System for AutoML of Gradient Boosted Trees}


\author[1]{Aleksei Sorokin\fnref{fn1}}
\ead{asorokin@hawk.iit.edu}

\author[2]{Xinran Zhu\fnref{fn1}}
\ead{xz584@cornell.edu}

\author[3]{Eric Hans Lee\corref{cor1}}
\ead{eric.lee@intel.com}

\author[3]{Bolong Cheng}
\ead{harvey.cheng@intel.com}

\affiliation[1]{organization={Illinois Institute of Technology},
}
\affiliation[2]{organization={Cornell University},
}
\affiliation[3]{organization={SigOpt: An Intel Company},
}

\fntext[fn1]{Both authors contributed equally to this research. Work done while at SigOpt.}
\cortext[cor1]{Corresponding author}

\begin{abstract}
Gradient boosted trees (GBTs) are ubiquitous models used by researchers, machine learning (ML) practitioners, and data scientists because 
of their robust performance, interpretable behavior, and ease-of-use. 
One critical challenge in training GBTs is the tuning of their hyperparameters. In practice, selecting these hyperparameters is often done manually.  
Recently, the ML community has advocated for tuning hyperparameters through black-box optimization and developed state-of-the-art systems to do so. 
However, applying such systems to tune GBTs suffers from two drawbacks. 
First, these systems are not \textit{model-aware}, rather they are designed to apply to a \textit{generic} model; this leaves significant optimization performance on the table. 
Second, using these systems requires \textit{domain knowledge} such as the choice of hyperparameter search space, which is an antithesis to the automatic experimentation that black-box optimization aims to provide. 
In this paper, we present SigOpt Mulch, a model-aware hyperparameter tuning system specifically designed for automated tuning of GBTs that provides two improvements over existing systems. 
First, Mulch leverages powerful techniques in metalearning and multifidelity optimization to perform model-aware hyperparameter optimization.
Second, it automates the process of learning performant hyperparameters by making intelligent decisions about the optimization search space, thus reducing the need for user domain knowledge. 
These innovations allow Mulch to identify good GBT hyperparameters far more efficiently ---and in a more seamless and user-friendly way--- than existing black-box hyperparameter tuning systems.
\end{abstract}




\end{frontmatter}



\section{Introduction}\label{sec:introduction}
In recent years, gradient boosted trees (GBTs) have become some of the most popular machine learning (ML) models for tasks such as classification, ranking, or regression. 
GBT software such as XGBoost \cite{chen2016xgboost}, LightGBM \cite{ke2017lightgbm}, or CatBoost \cite{prokhorenkova2018catboost} demonstrate strong performance in production ML pipelines, scientific applications, Kaggle competitions, and OpenML challenges \cite{uber-xgboost-blog, kaggle_survey, zhang2018data, zhong2018xgbfemf}. Furthermore, despite advances in deep neural networks (NNs), research continues to reaffirm the strong performance of GBTs on structured, tabular data compared to NNs \cite{shwartz2021tabular, qin2021neural}. 

Hyperparameter tuning is a key challenge in training and deploying machine learning models.
GBT hyperparameters include the number of boosting trees, learning rate, tree depth, and other parameters that trade off bias and variance of the model. Strategically tuning these hyperparameters often results in significant performance gains over default configurations. 
Unfortunately, GBT hyperparameter tuning is a difficult task. 
This is echoed in the XGBoost documentation, which laments that ``\textit{[hyper]parameter tuning is a dark art... it is impossible to create a comprehensive guide for doing so}'' \cite{chen2016xgboost}. 
As a result, tuning is typically performed by a domain expert who either leverages their intuition to manually tune the model ---or more often than not, simply picks the best from a randomly chosen selection \cite{bouthillier2020survey}. 
This resulting human-in-the-loop process is often inefficient and unprincipled. 

To address these challenges, the ML community has developed sophisticated systems to programmatically tune hyperparameters. 
Examples include open source packages such as Hyperopt \cite{bergstra2013hyperopt}, Scikit-Optimize \cite{markov2017skopt}, Optuna \cite{akiba2019optuna},  Ax/BoTorch \cite{balandat2020botorch}, and  Google Vizier \cite{golovin2017vizier}, as well as commercial solutions from SigOpt \cite{sigopt-documentation}, SAS Autotune \cite{koch2018autotune}, and Amazon SageMaker \cite{perrone2021sagemaker}. 
By modeling interactions among hyperparameters and making decisions accordingly, a technique that falls under \textit{black-box optimization}, these systems are usually able to identify more performant hyperparameters than hand-tuning within a small number of iterations. 
Black-box optimization makes minimal assumptions about the optimization problem, and only requires the ability to evaluate the function value (in contrast to gradient-based techniques, which also require the ability to evaluate the gradient). This makes it broadly applicable to a wide range of tasks, from searching NN architectures for computer vision to optimizing materials design parameters \citep{shahriari2016botutorial, malkomes2021cas, garnett_bayesoptbook_2022}.

While one can use existing systems to tune GBT hyperparameters, there are two drawbacks to using such generic optimization software for this specific task. 
First, black-box optimization is designed to make minimal assumptions about the objective function it is optimizing. 
However, in many modern ML systems, one knows \textit{exactly} which model they are training and what data they are training it with. 
These systems also repeatedly perform many of the same or similar training tasks. 
By both ignoring the structure of the problem and the data accrued from prior training tasks, one leaves performance on the table. 
Second, black-box optimization still requires the user to select the hyperparameters to optimize and the search space over which to optimize them; one must often exercise careful judgement to make informed choices about these decisions in order to achieve strong optimization performance. 
This prevents non-expert users, who may not possess such information, from unlocking the full potential of GBTs.

In this paper, we describe SigOpt Mulch, a system specifically designed for tuning GBTs. 
By circumventing the black-box approach, Mulch presents two corresponding advantages over standard hyperparameter tuning systems: 
\begin{itemize}
    \item \textbf{Performance:} Using \textit{data-driven} and \textit{model-aware} methods in \textit{metalearning} and \textit{multifidelity optimization}, respectively, Mulch is able to reliably identify better GBT hyperparameters in less time compared to open source HPO software. 
    \item \textbf{Automation:} Mulch attempts to automate GBT hyperparameter tuning as much as possible by metalearning intelligent defaults and providing a simplified API, thus relaxing the domain expertise required to perform HPO.
\end{itemize}
Mulch is a production system, so we communicate both the user-facing product and its underlying algorithmic contributions in a cohesive manner. As such, this paper is structured as follows. 
In Section \ref{sec:background}, we outline the general framework of hyperparameter tuning via black-box optimization. 
In \secref{sec:systems_api}, we describe the system architecture and the API design for Mulch and demonstrate its benefit over existing black-box optimization packages from a user perspective. 
In Sections~\ref{sec:metalearning} and \ref{sec:multifidelity}, we detail the algorithmic techniques employed to build our model-aware optimizer.
We demonstrate the improvement of Mulch over existing black-box optimization software
for tuning XGBoost in Section \ref{sec:ES}. 
Lastly, we provide some concluding thoughts in Section \ref{sec:conclusion}, as well as an analysis of Mulch's limitations and future work that might address these limitations.


\section{Background}
\label{sec:background}
\subsection{Hyperparameter Optimization}
\label{sec:hyperparameter_optimization}
Hyperparameter optimization (HPO) poses hyperparameter tuning as an optimization problem, 
$\min_{\ttheta \in \Theta} f(\ttheta)$, where $f$ is the objective function and $\Theta$ is the hyperparameter domain over which to search. In the context of GBTs, $f(\ttheta)$ is typically a metric quantifying classification or regression error, and $\Theta$ is a search space that contains GBT hyperparameters over which to search. Given a training and validation set, an example of this optimization problem is to minimize the root-mean-square-error (RMSE) on the validation set over GBT models with learning rate in $[0.1, 1]$ and number of estimators in $[10, 100]$. Generally speaking, HPO is expensive because each iteration requires fully training a model. Furthermore, HPO  often assumes that $f$ is a black-box function, meaning only the output, e.g., RMSE on the validation set, is known for any given input, e.g,. a hyperparameter configuration (so that no additional information such as gradients are available during the optimization procedure). The combination of expensive function evaluations and a lack of gradient information results in a very difficult optimization problem. 

In this paper, we utilize \textit{Bayesian optimization} (BO), a class of popular HPO methods that performs well in practice \cite{shahriari2016botutorial, turner2021bayesian} and is used in almost all existing HPO systems. 
BO is an iterative optimization method. 
At each iteration, it builds a probabilistic model that correlates the objective function with hyperparameter configurations. 
It then uses this model to determine the next hyperparameter configuration by optimizing a utility function called the \textit{acquisition function}. We note that BO must be started with an exploration phase that distributes a few points evenly throughout the search space, which act as initial training data for the probabilistic model. 

The most common probabilistic model used in BO is the \textit{Gaussian process} (GP) \cite{rasmussen_gpforml}. Given a set of hyperparameters $\ttheta_1, \dots, \ttheta_n$, a GP model correlates their $f$ values using a mean function and covariance kernel, which control how well the GP models the objective $f(\ttheta)$. Crucially, these functions themselves possess parameters. Typical parameters include kernel lengthscales, which control the multi-modality of the GP along each dimension. Shorter lengthscales imply less smooth local behavior ---that is, a GP with more local minima--- and vice versa. Calibrating these kernel parameters to accurately model the underlying objective function is important to achieve fast BO convergence. In practice, the kernel parameters are learned through standard statistical inference methods such as maximum likelihood estimation (MLE) \cite{rasmussen_gpforml, frazier2018tutorial}.

There is a fundamental trade-off between the dimensionality of the hyperparameter search space and effectiveness of HPO. A low-dimensional search space is easy to optimize, but could restrict the expressiveness of models found inside that space ---HPO inside this space may yield a poor model. Conversely, a high-dimensional space allows for expressive models, but is difficult to optimize ---HPO inside this space may not be able to identify a performant model in a reasonable amount of time. Balancing the size of the search space and the quality of model found through HPO is a difficult problem in practice, and is often left up to the practitioner's experience and preference.

\subsection{Hyperparameter Software Packages}
This section surveys some of the more popular software packages for hyperparameter optimization. 

\vspace{3mm} \noindent
\textbf{Optuna:}
Optuna \cite{akiba2019optuna} is likely one of the most popular open source packages for hyperparameter optimization due to its ease of use. It offers a define-by-run API that allows the user to build expressive and dynamic search spaces, built-in early stopping, and uses a variant of Bayesian optimization under the hood. What Optuna gains in ease of use it loses in flexibility; its underlying optimization algorithms are not easy to modify. 

\vspace{3mm} \noindent
\textbf{Scikit-optimize:}
Scikit-optimize \cite{scikit-learn} aims to be a lightweight and simple optimization package that uses Bayesian optimization under the hood. It is not as comprehensive as other hyperparameter optimization packages, in the sense of offering a wide variety of functionalities, but is popular due to being built directly with the ubiquitous \texttt{scikit-learn} package.

\vspace{3mm} \noindent
\textbf{Hyperopt:}
Hyperopt \cite{bergstra2013hyperopt} is another optimization package offering Bayesian optimization in a straightforward and lightweight API. It also offers integrations with Spark and MongoDB \cite{zaharia2010spark, banker2016mongodb}.


\subsection{GBTs and Their Hyperparameters}
Here we provide only a brief overview of gradient boosted trees (GBTs) \cite{mason1999boosting}.
A GBT is a set of decision trees, which are trained in sequence and predict the label of a data instance
as the sum of its decision trees. At each iteration, a GBT iteratively learns the decision trees by fitting the residual error of a loss function which generally contains an error term (for accuracy) and a regularization term (for model complexity). 

We represent the set of GBT hyperparameters by the vector $\ttheta$, which contains all integer, continuous, discrete, and categorical parameters that govern the GBT training process but cannot be learned through training itself. 
Examples include the learning rate, the number of estimators, or even the class of learning algorithm used. 
Different implementations of GBTs possess somewhat different hyperparameter sets, albeit with significant overlap. 

Regardless of implementation, the underlying problem remains the same; how to efficiently extract a performant configuration from the large space of possible hyperparameters. For example, there are over 30 XGBoost hyperparameters \cite{chen2016xgboost}. 
A challenge of GBT hyperparameter tuning is determining which hyperparameters to tune, the search space over which to tune them, and the number of trials with which to tune them.
In \secref{sec:metalearning}, we discuss how to automatically make these decisions for GBTs.


\section{Systems and API Description}
\label{sec:systems_api}

\subsection{System Design}

\begin{figure*}[ht!]
    \centering
     \begin{subfigure}[t]{0.43 \textwidth}
         \centering
         \includegraphics[width=\textwidth]{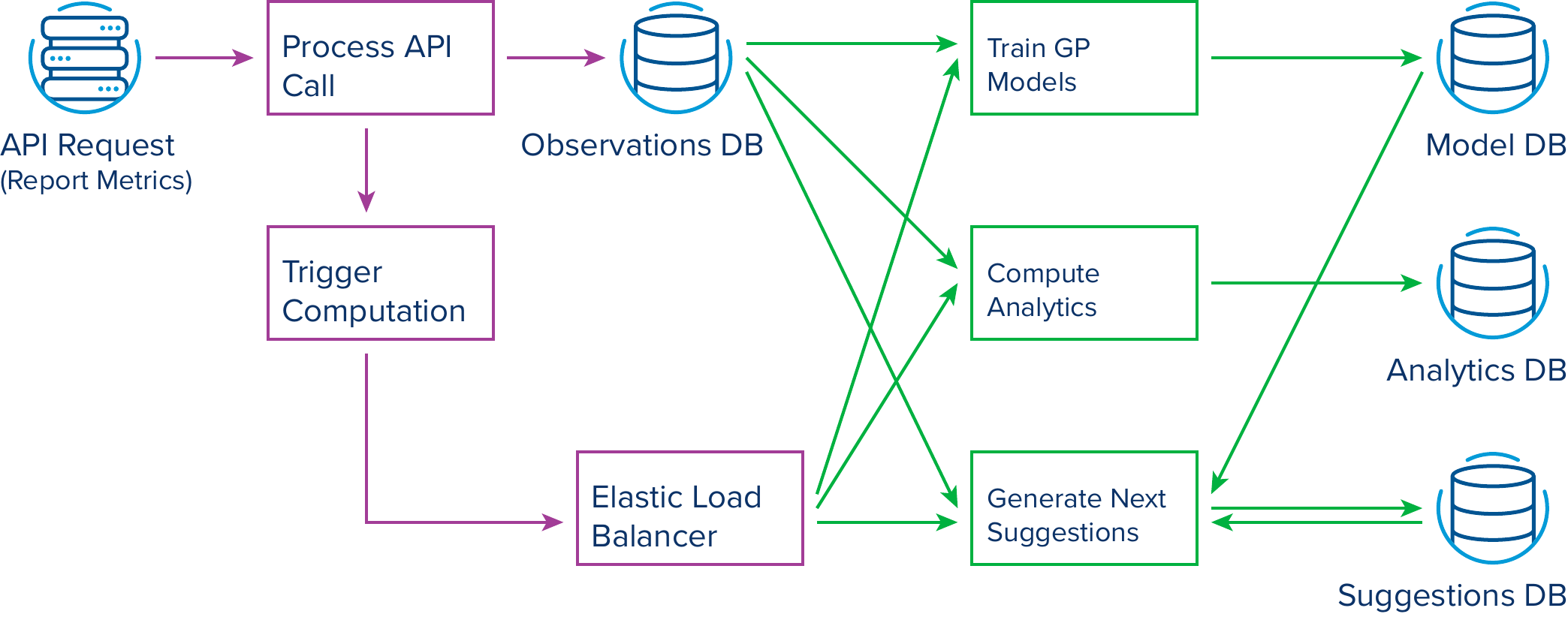}
         \caption{Asynchronous computing workflow.}
         \label{fig:asynchronous_compute}
     \end{subfigure}
     \hspace{7mm}
     \begin{subfigure}[t]{0.43 \textwidth}
         \centering
         \includegraphics[width=\textwidth]{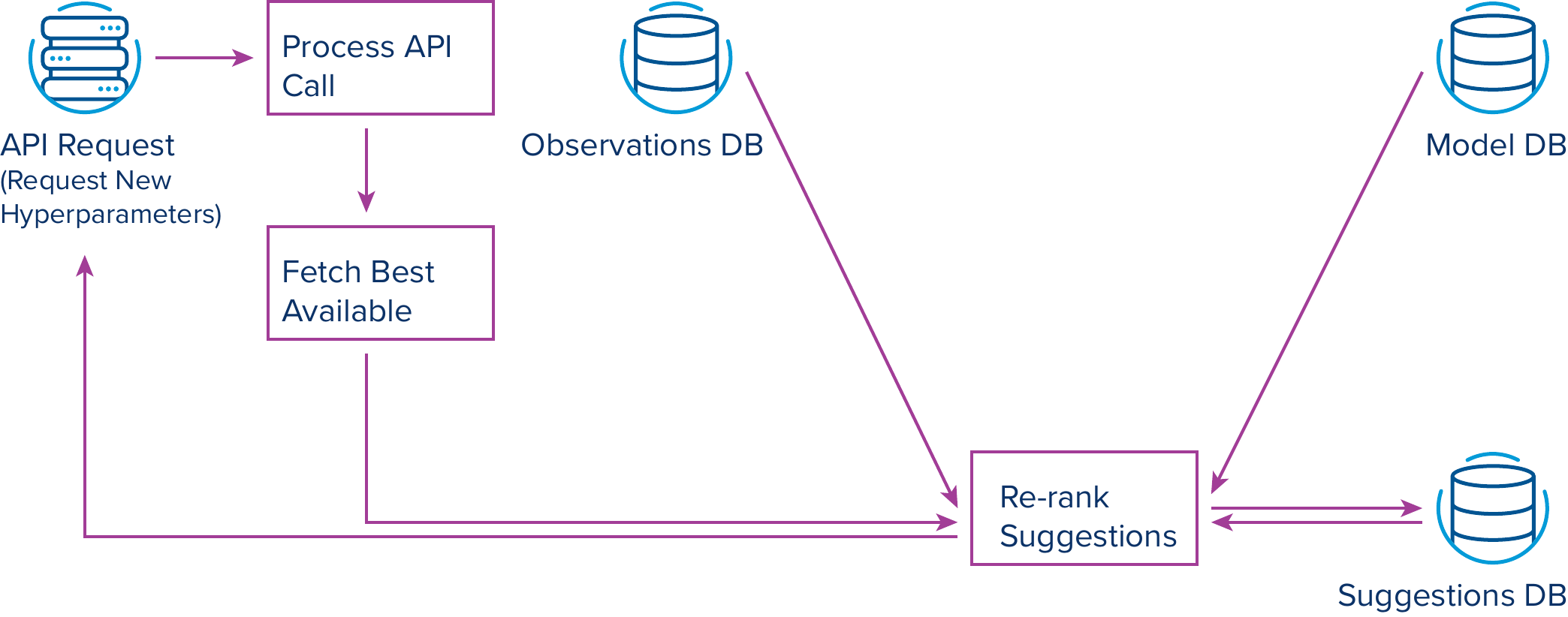}
         \caption{Synchronous computing workflow.}
         \label{fig:synchronous_compute}
     \end{subfigure}
    \caption{Overview of SigOpt's HPO computation workflows. All elements labeled in purple are synchronous computing processes. All elements labeled in green are asynchronous computing processes.}
    \label{fig:system_overview}
\end{figure*}

Mulch is a product built on top of SigOpt \cite{sigopt-documentation}, a cloud-based black-box optimization software. 
Mulch is specifically designed for efficiently optimizing GBT models. In the following subsections, we first overview SigOpt's system architecture and the vanilla SigOpt API. We then outline the new Mulch API. For the rest of this paper, we focus on tuning XGBoost \cite{chen2016xgboost}, but the resulting design principles and analysis apply to other GBT libraries. 

SigOpt is a hosted service of general black-box optimization tools built with the following considerations in mind:
\begin{itemize}
\item \textbf{Ease:} A lightweight RESTful API that requires only nominal user configuration, and an interactive and intuitive dashboard for analyzing HPO results.
\item \textbf{Flexibility:} Pause, continue, or modify the underlying optimization problem at any point of the HPO process.
\item \textbf{Availability:} Guaranteed, real-time hyperparameter configuration suggestions.
\item \textbf{Extensibility:} Easy to update and improve optimization algorithms in the backend.
\item \textbf{Scalability:} Handle a variety of use cases, from tuning time-consuming neural networks to optimizing simulation models with many parallel workers.
\end{itemize}

When users create an HPO job (called an \textbf{experiment}) using SigOpt, they only need to specify the parameter types and bounds, and the name of the relevant optimization metrics. 
The users train and evaluate the models on their own compute infrastructure and only interact with SigOpt via a RESTful API.
Users then iteratively request a new hyperparameter configuration and report back the evaluation metrics of the trained model (called a \textbf{run}) \cite{sigopt-documentation}. 
SigOpt provides a web dashboard for tracking the progress of an experiment. Users can interact with the dashboard to analyze the results of an experiment via multiple interactive plots, 
and they can modify the state of an experiment such as changing the parameter bounds or optimization budget. 
Like most popular black-box optimization software, the base SigOpt optimization system makes minimal assumptions about the problem that it is optimizing.

SigOpt is employed in different domains with vastly different requirements for the optimization engine. For example, a user tuning a computer vision model might only request a new hyperparameter every few hours with a total tuning budget of 50 runs due to the model's long training time. In contrast, another user optimizing a numerical simulation model might request batches of 100 suggestions every minute for a total budget of 5000 due to their highly specialized parallel compute infrastructure. To meet different criteria, SigOpt decouples the standard Bayesian optimization loop (as described in \secref{sec:hyperparameter_optimization}) into modular components to ensure that the
API response time does not exceed a few hundred milliseconds. 

In the \figref{fig:system_overview}, we show a high level diagram of how this decoupled BO process works. Whenever the user reports metric values back (\figref{fig:asynchronous_compute}), this triggers a series of computations such as refitting the lengthscales of the GPs and computing new suggestions using the updated information. These time-consuming computations are done asynchronously server-side and their results are stored in persistent databases. When the user requests new hyperparameter suggestions (\figref{fig:synchronous_compute}), this triggers a synchronous computation worker to fetch the existing suggestions stored in the database and re-rank them according to the most up-to-date GP model and acquisition function. To ensure that the suggestion database is always populated, SigOpt uses a fall-back suggestion generation mechanism that is computationally cheap, such as drawing samples from a predetermined distribution.

Since each block of the BO computation is decoupled, SigOpt can select different BO algorithms based on the experiment properties, similar to processes described in Google Vizier \cite{golovin2017vizier}. These properties may specify multiple objectives, constraints (in parameter space, metric space, or both), minimum performance thresholds, or any combination of these and more. Each combination of these may require a different solution strategy and thus a different acquisition function. Crucially, SigOpt accounts for this by automatically switching the acquisition functions without user intervention. This stands in contrast to open-source software, whose user must possess extensive knowledge of BO in order to select a suitable acquisition function for the optimization task at hand.  

\subsection{SigOpt API}

\begin{figure}[ht]
    

    
    
\caption{An HPO example of tuning XGBoost using the vanilla SigOpt API. The user needs to specify the search space (variable type and bounds), define the optimization metric (including code to compute it), and provide an optimization budget. Note that regardless of the system used for hyperparameter optimization, the user will always need to define these things before optimization occurs. }
\label{fig:sigopt_api}
\end{figure}
\begin{figure}[ht]
        
    
    
\caption{An HPO example of tuning XGBoost using the Mulch API. This code has the same functionality as \figref{fig:sigopt_api} but is significantly shorter. Note that it is significantly simpler to use; Mulch will automatically select the search space, optimization metric, and optimization budget. This automation loosens the domain expertise required to perform hyperparameter optimization. Mulch also allows an expert to make these decisions should they choose to do so.}
\label{fig:mulch_api}
\end{figure}

To create a standard SigOpt experiment, a user needs to define the parameters, their associated type and bounds (distinct elements in the case of categorical parameters), the metrics, and the tuning budget.
In \figref{fig:sigopt_api}, we demonstrate using the vanilla SigOpt API to create an experiment tuning XGBoost models.

The vanilla SigOpt API is designed to be lightweight in order to work with any type of machine learning model; therefore it places responsibility on the user to define a reasonable search space and pick the appropriate optimizing metrics\footnote{Virtually all open-source and commercial systems share the same design philosophy \citep{bergstra2013hyperopt, markov2017skopt, sigopt-documentation, golovin2017vizier, koch2018autotune, perrone2021sagemaker}.}. These decisions are barriers for newer users; knowing which hyperparameters to tune and the appropriate search space is not a trivial task and making these decisions may require years of experience.

\subsection{Mulch API and System Design}

\noindent
Mulch offers a streamlined API because it knows the model it is tuning. In \figref{fig:mulch_api}, we show an example of this updated API that has comparable functionality to \figref{fig:sigopt_api}. 
The \texttt{sigopt.xgboost.experiment} function completely abstracts away the experiment creation (lines 25-36), the model training (lines 8-22), and the BO loop (lines 39-41) in the vanilla SigOpt API code. 
Unlike vanilla SigOpt and other similar systems, Mulch shoulders the burden of setting up the HPO job such as determining the hyperparameter search space. 
These intelligent decisions make Mulch more \textit{accessible to a broader audience}, who may not have as much experience with either HPO or GBTs. 
Note that in this new API, the parameters (search space), metric, and budget definition are optional inputs (of the \texttt{experiment\_config} dictionary) and can always be specified if experienced XGBoost users want to maintain control over the HPO process.

Mulch takes SigOpt's automatic algorithm switching to the logical next step by also accounting for the model itself, allowing Mulch's backend to leverage more efficient model-aware optimization algorithms, which we detail in Sections \ref{sec:metalearning} and \ref{sec:multifidelity}. 
Finally, Mulch natively enables recording of each XGBoost model's metadata, learning curve, and additional metrics without additional code that vanilla SigOpt requires. 
We provide a detailed description of Mulch's model tracking capabilities in \appref{sec:appendix_mulch_model_tracking}.

\section{Learning Tasks}\label{sec:datasets_and_hyperparameters}

Developing Mulch required that we first build a suite of hyperparameter optimization problems. We refer to these problems as learning tasks to be consistent with the AutoML literature \cite{feurer2019auto}. To be more precise, a learning task consists of a dataset, an objective function to optimize, and a hyperparameter search space. 

In this section, we summarize the learning tasks we consider, which all involve learning the optimal hyperparameters of gradient boosted trees.  The overall goal of Mulch is to perform better on these tasks than existing software systems in production today with the hope that the resulting performance will generalize to unseen tasks. Consequently, we very carefully curated these tasks to represent difficult, real-world scenarios that a user will encounter, and we describe the method and rational for our curation below. 
\subsection{Datasets}
\begin{table*}[t!]
    \centering
    \small
       \caption{The datasets used in this paper represent a wide range of classification tasks of varying size and difficulty. For datasets labeled \highlighttwo{\textit{metalearning}}, we generate $2^{10}$ random hyperparameters over a 12d search space detailed in \tabref{tab:full_search_space}, which we use to learn an improved HPO procedure. Datasets labeled \highlightone{\textit{performance testing}} are, as suggested, used to test the performance of our improved HPO procedure. }
    \begin{tabular}{r|c|c|c|l}
        \toprule
        \textbf{Dataset Name} & \textbf{Dataset Size} & \textbf{Dimension} & \textbf{\# Classes} & \textbf{Used For}\\ \midrule
        \highlighttwo{\texttt{adult}} & 48,842 & 14 & 2 & \highlighttwo{\textit{metalearning}} \\ 
        \highlighttwo{\texttt{australian}} & 690 & 14 & 2 & \highlighttwo{\textit{metalearning}} \\ 
        \highlightone{\texttt{bands}} & 512 & 39 & 2 & \highlightone{\textit{performance testing}} \\ 
        \highlighttwo{\texttt{biodegredation}} & 1,055 & 41 & 2 & \highlighttwo{\textit{metalearning}} \\ 
        \highlighttwo{\texttt{breast-cancer}} & 286 & 9 & 2 & \highlighttwo{\textit{metalearning}} \\ 
        \highlighttwo{\texttt{car}} & 1,728 & 6 & 4 & \highlighttwo{\textit{metalearning}} \\ 
        \highlightone{\texttt{cmc}} & 1,473 & 9 & 3 & \highlightone{\textit{performance testing}} \\ 
        \highlighttwo{\texttt{default-credit}} & 30,000 & 24 & 2 & \highlighttwo{\textit{metalearning}} \\ 
        \highlighttwo{\texttt{dermatology}} & 366 & 33 & 6 & \highlighttwo{\textit{metalearning}} \\ 
        \highlighttwo{\texttt{diagnosis}} & 58,509 & 49 & 11 & \highlighttwo{\textit{metalearning}} \\ 
        \highlighttwo{\texttt{dow-jones-index}} & 750 & 16 & 2 & \highlighttwo{\textit{metalearning}} \\ 
        \highlighttwo{\texttt{eeg-eye-state}} & 14,980 & 15 & 2 & \highlighttwo{\textit{metalearning}} \\ 
        \highlighttwo{\texttt{firm-teacher}} & 10,800 & 20 & 4 & \highlighttwo{\textit{metalearning}} \\ 
        \highlightone{\texttt{flag}} & 194 & 30 & 6 & \highlightone{\textit{performance testing}} \\ 
        \highlighttwo{\texttt{forest-types}} & 523 & 27 & 4 & \highlighttwo{\textit{metalearning}} \\ 
        \highlighttwo{\texttt{german-numeric}} & 1,000 & 20 & 2 & \highlighttwo{\textit{metalearning}} \\ 
        \highlightone{\texttt{guillermo}}   &    20,000  & 4,297  &  2 & \highlightone{\textit{metalearning}} \\
        \highlighttwo{\texttt{haberman}} & 306 & 3 & 2 & \highlighttwo{\textit{metalearning}} \\ 
        \highlightone{\texttt{hill-valley}} & 1,210 & 101 & 2 & \highlightone{\textit{performance testing}} \\ 
        \highlighttwo{\texttt{iris}} & 150 & 4 & 3 & \highlighttwo{\textit{metalearning}} \\ 
        \highlighttwo{\texttt{magic04}} & 19,020 & 11 & 2 & \highlighttwo{\textit{metalearning}} \\ 
        \highlighttwo{\texttt{mammographic-masses}} & 961 & 6 & 2 & \highlighttwo{\textit{metalearning}} \\ 
        \highlighttwo{\texttt{parkinsons}} & 197 & 23 & 2 & \highlighttwo{\textit{metalearning}} \\ 
        \highlightone{\texttt{pima-indians-diabetes}} & 768 & 8 & 2 & \highlightone{\textit{performance testing}} \\ 
        \highlighttwo{\texttt{real-sim}}   & 72,309 &  20,958  &2 & \highlighttwo{\textit{metalearning}} \\
        \highlightone{\texttt{relax}} & 182 & 13 & 2 & \highlightone{\textit{performance testing}} \\ 
        \highlightone{\texttt{rcv1}} & 677,399 &  47,236  &2 & \highlightone{\textit{performance testing}} \\
        \highlighttwo{\texttt{satellite}} & 6,435 & 36 & 6 & \highlighttwo{\textit{metalearning}} \\ 
        \highlightone{\texttt{sonar}} & 208 & 60 & 2 & \highlightone{\textit{performance testing}} \\ 
        \highlighttwo{\texttt{spambase}} & 4,601 & 57 & 2 & \highlighttwo{\textit{metalearning}} \\ 
        \highlightone{\texttt{tae}} & 151 & 5 & 3 & \highlightone{\textit{performance testing}} \\ 
        \highlighttwo{\texttt{tic-tac-toe}} & 958 & 9 & 2 & \highlighttwo{\textit{metalearning}} \\ 
        \highlighttwo{\texttt{transfusion}} & 748 & 5 & 2 & \highlighttwo{\textit{metalearning}} \\
        \highlighttwo{\texttt{usps}}   & 9,298   &  256     &10 & \highlighttwo{\textit{metalearning}} \\
        \highlightone{\texttt{vertebrael3}} & 310 & 6 & 3 & \highlightone{\textit{performance testing}} \\ 
        \highlighttwo{\texttt{waveformnoise}} & 5,000 & 40 & 3 & \highlighttwo{\textit{metalearning}} \\ 
        \highlighttwo{\texttt{wdbc}} & 569 & 32 & 2 & \highlighttwo{\textit{metalearning}} \\ 
        \highlighttwo{\texttt{wholesale}} & 440 & 8 & 2 & \highlighttwo{\textit{metalearning}} \\ 
        \highlightone{\texttt{wine}} & 178 & 13 & 3 & \highlightone{\textit{performance testing}} \\ 
        \highlightone{\texttt{yahoo-ltrc}} & 473,134 & 700 & 5 & \highlightone{\textit{performance testing}} \\ 
        \bottomrule
    \end{tabular}
    \label{tab:datasets}
\end{table*}

We selected 40 learning tasks drawn from the OpenML classification challenge \cite{bischl2017openml} and the AutoML classification benchmark \cite{gijsbers2019open}, which used both for analysis and empirical evaluation. These datasets have been curated from a larger body of about 85 learning tasks to emphasize the following two properties. 
\begin{itemize}
    \item \textbf{Balance:} Classes should be relatively balanced to avoid outlier detection learning tasks. In the case of high imbalance, one should weigh class instances and select an appropriate optimization metric. We consider this beyond the scope of this paper, and therefore removed all datasets with a large class imbalance (in particular, where one class is more than 80 percent of all training instances). 
    \item \textbf{Complexity:} Easy learning tasks where the majority of hyperparameters yield a good model do not provide useful metalearning data nor do they serve as useful benchmarks. We removed datasets for which HPO was able to consistently find a good model in ten iterations.
\end{itemize}
Generally speaking, these learning tasks are binary and multiclass classification over both sparse and dense datasets of size between a few kilobytes to a few gigabytes in size. 
After curating these datasets, we further subdivide them into learning and testing; 
see \tabref{tab:datasets} for a complete breakdown of the datasets, some of their properties, and whether they are used for learning or testing. 

\subsection{Hyperparameters}
\label{sec:hyperparameters}

\begin{figure*}[t]
    \centering
    \includegraphics[width=0.95\textwidth]{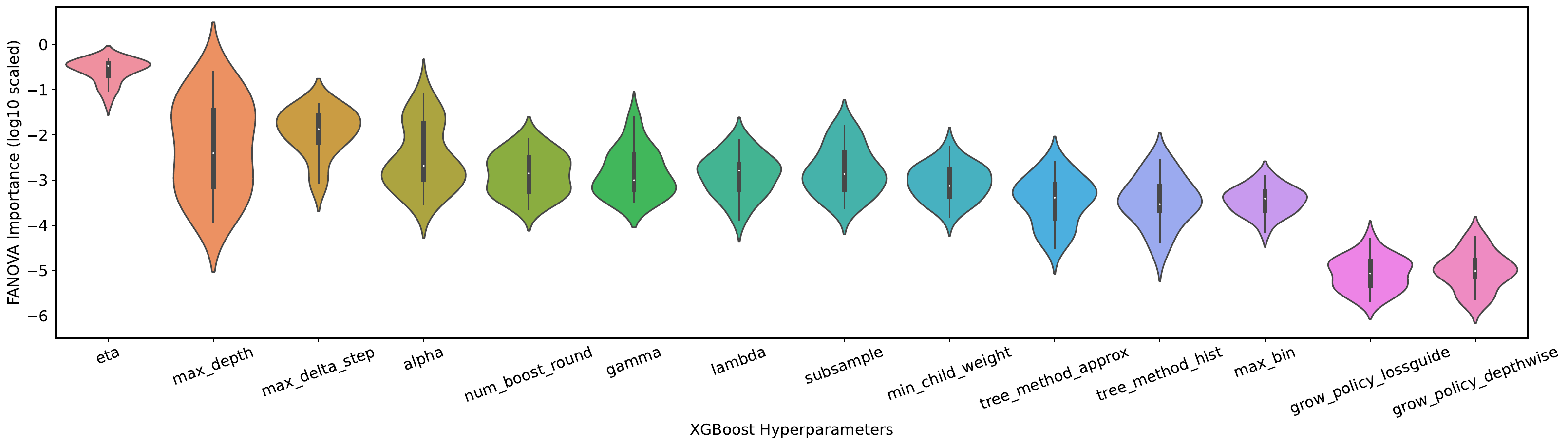}
    \caption{A violin plot of XGBoost hyperparameter importance scores sorted by mean importance score in decreasing order and log scaled. Each violin plots the distribution of a hyperparameter's importance scores across our suite of learning tasks. Perhaps unsurprisingly, the learning rate \texttt{eta} is the most important hyperparameter. We did not consider hyperparameters that affect XGBoost's alternative learning modes, such as boosted linear models or dropout trees. 
    }\label{fig:fanova_violin_expanded}
\end{figure*}

One of the key problems to answer, before proceeding with HPO, is to determine which hyperparameters to tune. As mentioned in \secref{sec:hyperparameter_optimization}, there is a fundamental trade-off between the dimensionality of the hyperparameter search space and the expressiveness of models found in the search space; a high-dimensional search space is too large to search efficiently, and a low-dimensional one may not contain a suitable model. Unfortunately, tuning in the full space of all design decisions is computationally prohibitive.

\begin{table*}[!ht]
    \centering
     \caption{The hyperparameter search spaces we considered, ordered in terms of importance and colored to be consistent with Figure \ref{fig:fanova_violin_expanded}} 
     \label{tab:full_search_space}
    \begin{tabular}{r|c|c| l}
        \toprule
         Hyperparamter Name & Hyperparameter Type & Search Domain  & Importance\\
         \midrule
         \texttt{eta}    & float        & $[-5,1]$ (\textit{log scaled})                  & \textcolor{Salmon}{ \hspace{6mm} Rank 1} \\ 
         \texttt{max\_depth}         & integer      & $[1,32]$                  & \textcolor{Peach}{ \hspace{6mm} Rank 2} \\
         \texttt{max\_delta\_step}   & float        & $[0,10]$                  & \textcolor{Tan}{ \hspace{6mm} Rank 3} \\
         \texttt{alpha}              & float        & $[0,10]$                  & \textcolor{GreenYellow}{ \hspace{6mm} Rank 4} \\
         \texttt{num\_boost\_round}  & integer      & $[1,500]$                 & \textcolor{YellowGreen}{ \hspace{6mm} Rank 5}  \\ 
         \texttt{gamma}              & float        &  $[0,5]$                  & \textcolor{Green}{ \hspace{6mm} Rank 6} \\
         \texttt{lambda}             & float        & $[0,10]$                  & \textcolor{SeaGreen}{ \hspace{6mm} Rank 7} \\
         \texttt{subsample}          & float        & $[0.5,1]$                 & \textcolor{BlueGreen}{ \hspace{6mm} Rank 8} \\
         \texttt{min\_child\_weight} & float        & $[1,5]$                   & \textcolor{Turquoise}{ \hspace{6mm} Rank 9} \\
         \texttt{tree\_method}       & categorical  & \{approx, hist\}          & \textcolor{Cyan}{ \hspace{6mm} Rank 10} \\
         \texttt{max\_bin}           & integer      & $[128,512]$               & \textcolor{Orchid}{ \hspace{6mm} Rank 11} \\
         \texttt{grow\_policy}       & categorical  & \{depthwise, lossguide\}  & \textcolor{Thistle}{ \hspace{6mm} Rank 12} \\
         \bottomrule
    \end{tabular}
\end{table*} 

The question becomes how to select an important subset of hyperparameters whose search space contains suitable models and is not prohibitively large. Finding these important hyperparameters enables more efficient optimization. 

We do so through functional analysis of variance (FANOVA), one of the standard methods of quantifying a hyperparameter's importance \cite{hutter2014efficient}. At a high level, FANOVA decomposes a function into additive, orthogonal sub-functions, each dependent only on a distinct subset of inputs. The variance of a sub-function indicates the variability in a metric, e.g. accuracy or RMSE, attributable to the dependent subset of hyperparameters. For individual hyperparameters, i.e. size $1$ subsets, normalizing the variance of the corresponding sub-functions by the variance of the full function results in \textit{individual importance scores}. Importance scores for higher order interactions, i.e. those subsets of size greater than $1$, require a generalized formulation, see \cite{owen2019monte} for details. 

To perform FANOVA, each of our 28 learning tasks (highlighted in orange in \tabref{tab:datasets}) is evaluated at $2^{10}$ quasi-random hyperparameter configurations in the full $12$-dimensional search domain provided in \tabref{tab:full_search_space}.
We use the \texttt{fanova} Python package \cite{hutter2014efficient} to compute the importance scores. We showcase the distribution of the importance scores in \figref{fig:fanova_violin_expanded} as a violin plot and rank them according to the mean importance scores. 
We caution the reader that these importance scores are not absolute; they are conditioned on the choice of search space, learning task, and optimization metric.
However, we believe their ordering is relatively invariant under reasonable perturbations of these decisions. 

These importance scores provide guidance for Mulch to select which hyperparameters to tune. We adopt the straightforward selection strategy of keeping the $d$ hyperparameters with largest individual importance scores. While more sophisticated strategies may be advantageous, we leave such investigations to future work.


\section{Metalearning}\label{sec:metalearning}
\begin{figure*}[ht]
    \centering
    \includegraphics[width=\textwidth]{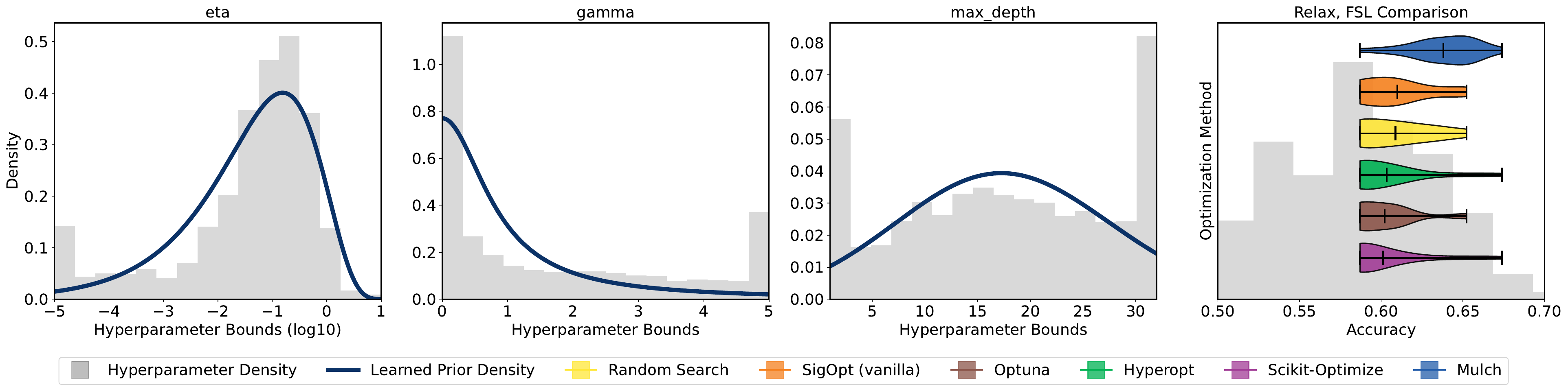}
    \caption{On the left three plots, we show some priors (dark blue) that have been fit to an empirical density (grey) of previously identified hyperparameters that demonstrate strong performance. In the right-most plot, we demonstrate the strength of FSL using these prior beliefs compared to vanilla SigOpt (orange), Optuna (brown), Hyperopt (green), Scikit-Optimize (purple), and random search (yellow) on a carefully chosen example.   \label{fig:densities}}
\end{figure*}

Metalearning by learning-to-learn is a paradigm that systematically observes how machine learning algorithms perform on a set of learning tasks, and then uses this experience  to learn new tasks much faster than otherwise possible \cite{ metalearning_vanschoren}. When applicable, metalearning can improve performance of machine learning algorithms by streamlining them in a data-driven way. 

Metalearning is made possible by HPO systems repeatedly performing the same or similar learning tasks ---by focusing on GBTs, we can accumulate experience from all prior HPO tasks. We seek to leverage this wealth of information to make intelligent, data-driven decisions. 
We employ \textit{few-shot learning} and \textit{model-aware optimization} to accelerate optimization on this search space. The combination of these techniques allow us to
quickly identify the high performing hyperparameters.

\subsection{Few-shot Learning} 
\label{sec:few_shot_learning}

In the context of HPO, few-shot learning (FSL) seeks to learn performant hyperparameter configurations given a limited budget of attempts. This limited budget is prohibitive and prevents most methods from progressing past the initial exploration of the hyperparameter search space (as described in \secref{sec:hyperparameter_optimization}). 
Indeed, there is simply no way to reliably identify good hyperparameters in a few iterations without additional information to inform the optimization process. 

Mulch does use additional information in the form of priors over the hyperparameter search space, which is a fundamental methodology in FSL. A sufficiently informed prior can  overcome the restrictions of a limited budget by heavily biasing a new learning task towards a certain set of hyperparameters. 

A key (and perhaps unsurprising) observation allows us to construct such priors: most GBT learning tasks share a common preference towards certain regions of the hyperparameter search space ---or colloquially: \textit{some hyperparameter configurations are much more likely to be high performing than others}. In \figref{fig:densities} we show the distribution of the top ten percent of all hyperparameter configurations, whose histograms we plot in grey. The values of key hyperparameters such as \texttt{eta}, \texttt{gamma}, and \texttt{max\_depth} tend to accumulate over specific values. Furthermore, these preferences are often tightly concentrated, leading to naturally strong priors that can greatly accelerate GBT hyperparameter tuning by guiding optimization towards promising hyperparameter regions. 

We formally define a prior as the density 
\[
\rho(\theta_1 , \theta_2 , \dots,  \theta_d) : \Theta \rightarrow \mathbb{R},\]
where $\theta_1, \dots, \theta_d$ are random variables encoding $d$ hyperparameters. We assume statistical independence between hyperparameters and decompose the prior density as
\[
\rho(\theta_1 , \theta_2 , \dots,  \theta_d) = \rho_1(\theta_1)\cdots \rho_d(\theta_d).
\]
The goal now becomes building the $d$ independent one-dimensional densities. 
We construct these densities through \textit{metalearning} by accumulating previous hyperparameters that demonstrated strong performance on previous HPO runs \cite{perrone2018scalable} and fitting a distribution to the resulting samples. To fit the distributions, we pick either \textit{Half-Cauchy}, \textit{Beta}, \textit{Gamma}, and \textit{Uniform}\footnote{Uniform in a subdomain of the search space.}, and use MLE to fit the density parameters. We visualize some of these learned priors in \figref{fig:densities}, which demonstrate that for hyperparameters such as learning rate, there is a clear and informative distribution to be learned. 

We note that the strength of a highly concentrated prior is simultaneously a weakness. The more concentrated a prior, the more vulnerable it is to mis-specification; a situation in which a prior concentrates around an unsuitable region of the search space for a given learning task. To guard against this in practice, we learn multiple priors and \textit{ensemble} them. Ensembling is a straightforward way of averaging different models for robustness. In Mulch, we ensemble priors by learning one from each family, averaging the priors, and sampling from this combined distribution. By sampling these ensemble densities, we hope to guard against MLE overfitting. 

Once we have priors, our few-shot learning procedure is straightforward: sample from the priors, use these to warm-start the Gaussian process model, and run a few iterations of Bayesian optimization. We compare our few-shot learning procedure to Optuna, Hyperopt, standard SigOpt, Scikit-Optimize, and random search in the right-most panel of \figref{fig:densities}. Our FSL procedure outperforms all baselines as measured by mean performance. This particular learning task is difficult for black-box optimization methods, which tend to identify rather poor hyperparameter configurations, as evidenced by the baselines' left-heavy tails. On the other hand, FSL consistently identifies good configurations, as evidenced by its right-heavy tail. In other words, it is also the most robust method.

Our FSL procedure can also be viewed as a lightweight approach to ``warm start'' the Bayesian optimization process \cite{feurer2014warmstart}.
Additionally, the FSL sampling procedure also serves as a fallback suggestion generation mechanism.
\subsection{Model-Aware Bayesian Optimization} 
\label{sec:model_calibration}
Model-aware Bayesian optimization can speed up the HPO process over the black-box approach by incorporating additional information of the problem \cite{bayesoptworkshop2016}. 
We explore a simple idea of adapting domain knowledge into the BO process when tuning gradient boosted trees; more specifically, we try to improve on fitting of the underlying GP surrogate model.

Recall from \secref{sec:hyperparameter_optimization} that GPs have their own parameters (e.g., lengthscales) that need to be learned. This is typically accomplished via MLE using multi-started quasi-Newton methods over a bounded lengthscale domain \cite{shahriari2016botutorial}.
In most software implementations, these domain bounds are determined using heuristics and often left intentionally loose in order to work well in a wide range of applications\footnote{For example, scikit-learn's \cite{scikit-learn} implementation of Mat\'{e}rn kernel  (which Scikit-Optimize uses) has a default lengthscale bounds of $[10^{-4}, 10^4]$, a generous eight orders of magnitude difference.}.
By knowing the hyperparameters that we are tuning and their likely behaviors, we can meta-learn priors over the lengthscale domain. 
We meta-learn lengthscale priors in the same fashion we learn search space priors in \secref{sec:few_shot_learning}: aggregate the lengthscales of the best GP models over our suite of learning tasks, and fit a distribution to this aggregated data. We can then use this learned prior in future GP modeling, the most straightforward of which is replacing the MLE with a maximum \textit{a posteriori} (MAP) estimate \cite{rasmussen_gpforml}. We find it simplest to use a uniform prior over a smaller bounding box: the MAP estimate of the GP parameters then corresponds to an MLE estimate inside the smaller bounding box. 

Finally, we combine our earlier few-shot learning procedure with the model-aware Bayesian optimization described in this subsection. First, we sample from our learned priors to generate a number of initial samples. We then use these samples to warm-start BO, which will learn it's internal lengthscale priors via a MAP estimate. The resulting HPO procedure very clearly outperforms software that exists today, and we discuss this more in \secref{sec:ES}

\begin{figure}[t!]
    \centering
    \includegraphics[width=0.48\textwidth]{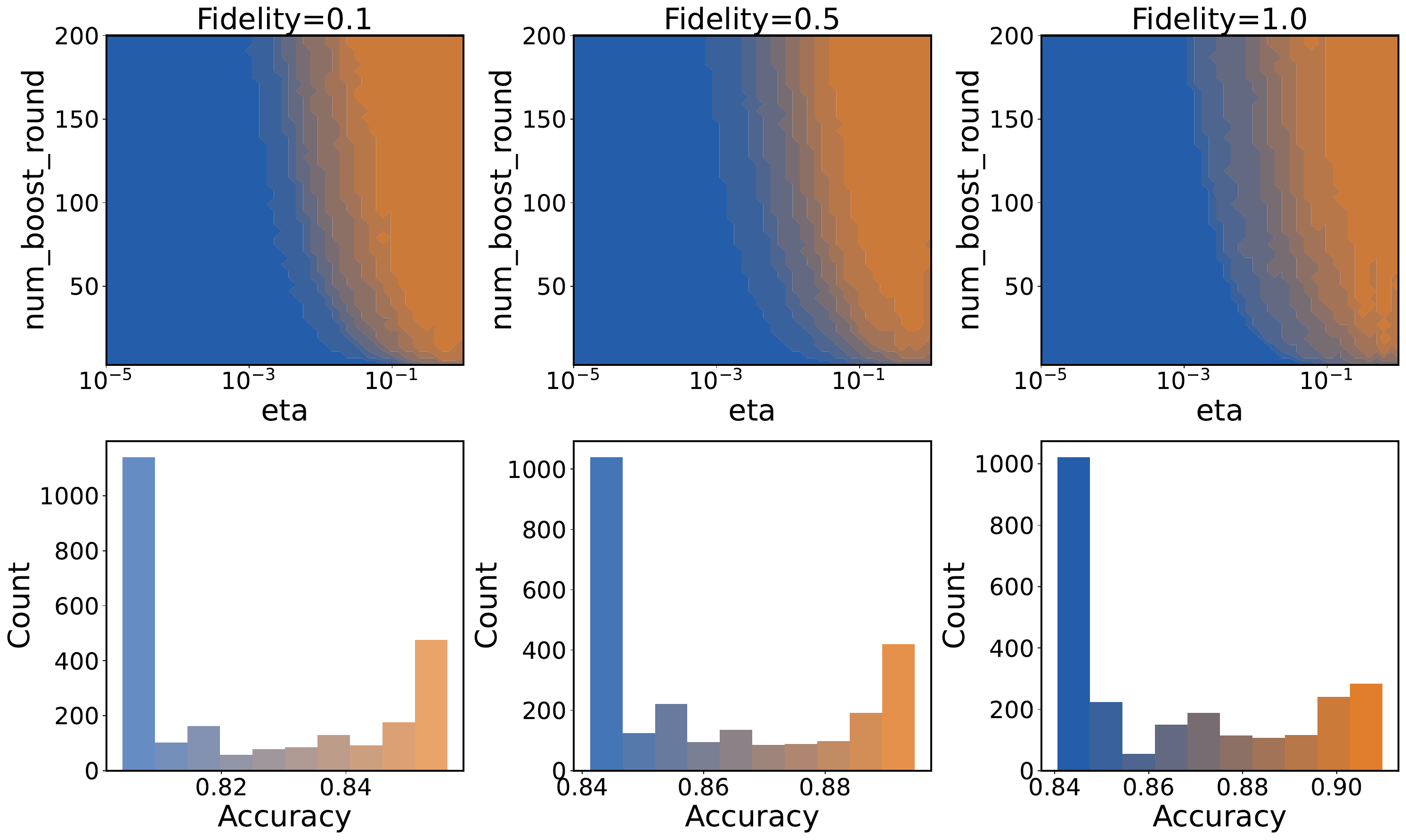}
    \caption{Contours and histograms of classification accuracy when tuning two XGBoost hyperparameters, \texttt{eta} and \texttt{num\_boost\_round}, on the Satellite dataset.  \label{fig: Figure_fidelity_study_2D_hist}}
\end{figure}


\section{Cost-Aware HPO}\label{sec:multifidelity}

In the previous section, we discussed how to be \textit{model-aware}; that is, to take into account the model in order to have Mulch make more intelligent optimization decisions. 
In this section, we discuss the how to be \textit{cost-aware}; that is, to take into account the cost of training during optimization. We do so through multifidelity optimization \citep{eggensperger2021hpobench, li2017hyperband, falkner2018bohb}, which is a well-known family of techniques to decrease the total time-to-model. 

\subsection{Multifidelity with Partial Data}
\label{sec:partial_data}

\begin{algorithm}[t]
\caption{Mulch-MF} 
\label{algo:Mulch-MF}
\textbf{Inputs}: Total budget $B$, the low-fidelity parameter $\rlow$, number of initial evaluations at low and high fidelity $n_l$ and $n_h$.
\begin{algorithmic}[1]
\small
    \State \textbf{Initialization:} Initialize $n_l$ and $n_h$ samples at the low fidelity and the high-fidelity respectively. Update total budget used. 
	\While{$b<B$}
		\State \textbf{Update GPs:} Update GP models $M_l$ and $M_h$, using all low-fidelity and high-fidelity evaluations, respectively.
		\State \textbf{Sample from GPs: } Based on acquisition function, select two samples $\ttheta_1$ and $\ttheta_2$ from $M_l$ and $M_h$, respectively.
		\State \textbf{Update sample pool $S$: } $$S = \{[\ttheta_1, C_l(\ttheta_1), C_h(\ttheta_1)], [\ttheta_2, C_l(\ttheta_2), C_h(\ttheta_2)]\},$$ Note that $C_l(\ttheta_1)+C_l(\ttheta_2) = C_h(\ttheta_1)+C_h(\ttheta_2) = 1$.
		\State \textbf{Select from $S$: } Select $\ttheta_l$ from $S$ based on $C_l(\ttheta_1)$, $C_l(\ttheta_2)$, and select $\ttheta_h$ from $S$ based on $C_h(\ttheta_1)$, $C_h(\ttheta_2)$. 
		\State \textbf{Evaluate selected samples: } evaluate $\ttheta_l$ at the low-fidelity $\rlow$ and evaluate $\ttheta_h$ at the high-fidelity $\rhigh=1$.
		\State $b = b + \rlow + \rhigh$.
	\EndWhile
\end{algorithmic}
\end{algorithm}

In optimization, often times the objective can be evaluated at multiple fidelities $f(\ttheta, r)$, where a fidelity parameter $r$ trades off the accuracy and cost of the evaluation. 
In this paper, we take $r$ to be the percentage of training data.
The training data size is proportional to the training time, and GBT models trained on partial data often perform comparably to models trained on full data.
One drawback of the existing multifidelity HPO algorithms is that they require the users to choose and define the number of fidelity levels and their respective costs \citep{eggensperger2021hpobench, li2017hyperband, falkner2018bohb}.
We want to automate this process with Mulch, which requires that we correlate XGBoost performance (accuracy) among different fidelity levels to facilitate the choice of fidelities.

To study the correlation between model performance and the fidelity level, we first perform a small-scale study in a 2D search space (\texttt{eta} and \texttt{num\_boost\_round}) on the Satellite dataset.
\figref{fig: Figure_fidelity_study_2D_hist} shows accuracy contours and histograms at different fidelity levels.
We observe that the contours exhibit the same shape across the different fidelity levels; good configurations (those yielding high accuracy) in full fidelity ($r=1.0$) are mostly well recognized in lower fidelity settings ($r=\{0.1, 0.5\}$).

To better quantify similarities between the lower fidelity models and the highest fidelity models, we define three scores. 
First, the \textit{correlation score} measures the similarity between low fidelity accuracy and full fidelity accuracy over the entire hyperparameter search space.
Second, the \textit{precision score} measures how likely good configurations in low fidelity (those with top $10\%$ accuracy) are also performant in full fidelity (sufficient condition). 
Lastly, the \textit{recall score} measures how likely good configurations in full fidelity are also performant in low fidelity (necessary condition).

To perform a large-scale analysis of these scores, we uniformly sample a 5D search space at fidelity levels $r = \{0.1, 0.3, 0.5, 0.7, 1.0\}$ over our suite of learning tasks and compute the average correlation, precision, and recall scores for each fidelity level.
From \tabref{table: fidelity study, score summary}, we observe correlation scores to be low and increase as $r$ increases, while precision and recall scores remain consistently high for all $r$'s, even when only using 10\% training data. 
The precision and recall scores are more important than correlation, since they reflect how well the low fidelity model can predict promising configurations.

\subsection{Multifidelity HPO}
\label{sec:mulch-mf}

Based on the multifidelity study, we develop a cost-aware HPO algorithm using two fidelity levels only ---one low fidelity using $\rlow$ proportion of the data, and the full fidelity $\rhigh=1$.
In addition to using cheaper low-fidelity evaluations to reduce HPO time, we also design the
algorithm to factor in the cost of each hyperparameter configuration. 
One notable example is \texttt{num\_boost\_round}, whose value is linearly proportional to training time.
In general, we want to evaluate expensive hyperparameter configurations, i.e., those with higher \texttt{num\_boost\_round} values, on the low-fidelity models.

Our algorithm, called Mulch-MF, builds two GP models: a GP $M_l$ built from all low-fidelity evaluations, and a GP $M_h$ built from all high-fidelity evaluations. 
At each iteration, each GP independently generates a hyperparameter configuration by maximizing the expected improvement acquisition function \cite{Jones1998ego}; we denote these as $\ttheta_1, \ttheta_2$ respectively.
For GBTs, we define the cost function $C_l(\cdot)$ to be proportional to the value of \texttt{num\_boost\_round} for the low fidelity model and $C_h(\cdot)$ to be inversely proportional to \texttt{num\_boost\_round} for the high fidelity model. 
We sample $\ttheta_l$ from $(\ttheta_1, \ttheta_2)$ according to the probability mass function $[C_l(\ttheta_1), C_l(\ttheta_2)]$ and then evaluate $\ttheta_l$ at fidelity $r_l$. 
In other words, $C_l(\cdot)$ is the probability that a sample $\ttheta$ will be selected for a low-fidelity evaluation. We do the same with $\ttheta_h$ and $C_h(\cdot)$. 
In this way, the two GPs interact and share information; the high-fidelity model $M_h$ has a larger chance to evaluate cheaper samples and vice versa. 
We present the pseudocode of Mulch-MF in Algorithm \ref{algo:Mulch-MF}.

\begin{table}[t]
\caption{Scores of different fidelity levels, computed from quasirandom samples of all XGBoost hyperparameters and averaged over our suite of learning tasks. \label{table: fidelity study, score summary}}
\begin{tabular}{l | r| r| r}
\toprule
$r$   & Correlation score & Precision score & Recall score \\
\midrule
0.1 & 0.523   & 0.869 & 0.823 \\
0.3 & 0.686   & 0.898 & 0.884  \\
0.5 & 0.705   & 0.901 & 0.897  \\
0.7 & 0.713   & 0.900 & 0.904 \\
\bottomrule
\end{tabular}

\end{table}
\subsection{Early Stopping}

GBT is an iterative learning process: at each step, learn a new decision tree to improve the model. 
Like with many iterative ML models, we might decide to terminate the learning process early if we do not see the model improving to save time \cite{prechelt1998early, zhang2005boosting}. The simplest form of early stopping terminates the learning process once predictive accuracy does not improve for a certain number of iterations $n_{s}$. Thus, $n_{s}$ trades off computational savings (the smaller $n_{s}$, the more aggressive the early stopping) with performance (aggressive early stopping might miss out on identifying good hyperparameters).

While many software and systems support early stopping, Mulch can natively enable early stopping for XGBoost without any additional user modification. This stands in contrast to black-box optimization software like Optuna, which requires the users to custom-write an additional stopping criterion. In practice, performance gains must be contextualized by the amount of effort spent to achieve said gains, whether it be human or compute. And Mulch's goal is for the users to expend as little effort as possible.


\section{Empirical Studies}\label{sec:ES}
\begin{figure*}[t]
    \includegraphics[trim={0 1.5cm 0 0},clip, width=\textwidth]{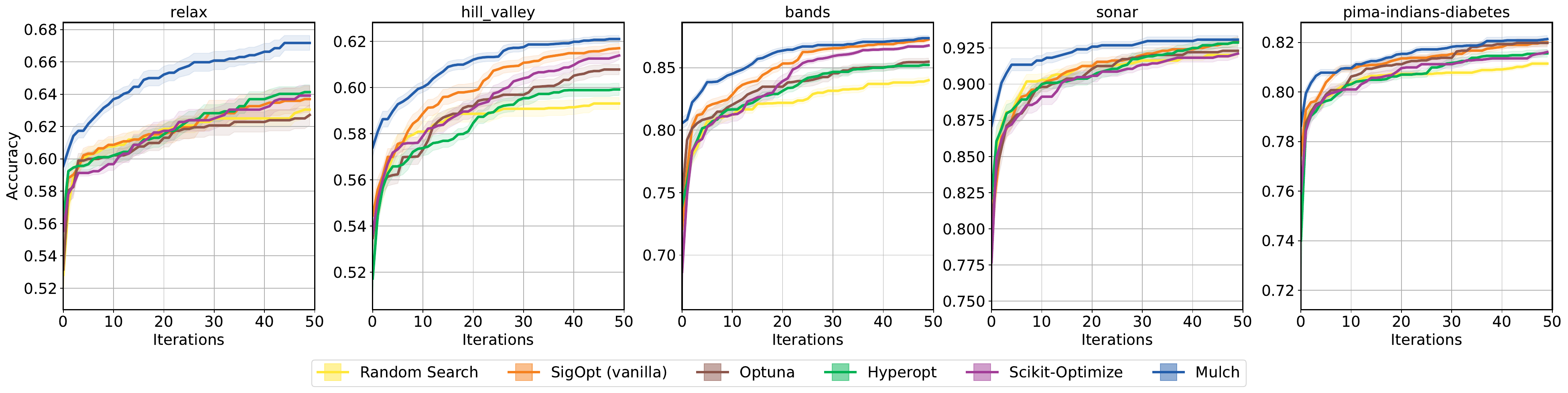}
    \includegraphics[width=\textwidth]{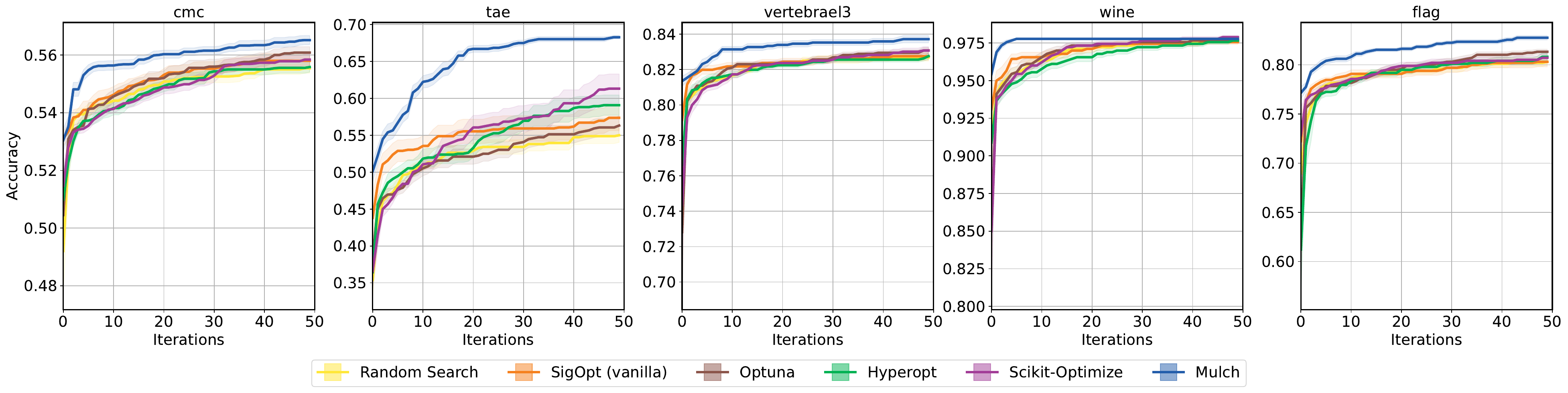}
    \centering
    \caption{Mulch's metalearning (blue) results in identification of better models compared to: vanilla SigOpt (orange), Optuna (black), Hyperopt (green), Scikit-optimize (purple), and random search (yellow). In this case, we benchmark Mulch on a set of 10 unseen classification problems and measure model performance through classification accuracy.}
    \label{fig:warmstart_full_bo_runs}
\end{figure*}

We compare Mulch to the following HPO software and systems: vanilla SigOpt (model agnostic), Hyperopt, Optuna, Scikit-Optimize, and random search. We choose to tune \texttt{eta}, \texttt{gamma}, \texttt{max\_depth}, \texttt{min\_child\_weight}, and \texttt{num\_boost\_round} with the search spaces in \tabref{tab:full_search_space}.

\subsection{Metalearning} \label{sec:meta_ES}
The first metalearning experiment is to use a straightforward few-shot learning scheme ---sampling eight times from our metalearned prior densities. This represents a situation in which standard HPO methods cannot effectively optimize and locate a good model within the restrictive optimization budget. As demonstrated in \figref{fig:densities} and \figref{fig:warmstart_full_bo_runs}, within the first eight iterations, Mulch rapidly outpaces all baselines on five unseen HPO tasks. We then use these FSL samples to initialize a standard BO routine within SigOpt. This enables Mulch to maintain a sizable performance advantage over the baselines, as seen again in \figref{fig:warmstart_full_bo_runs}. In fact, for these learning tasks, Mulch is more than twice as fast as the next runner-up. In 50 iterations, Mulch is able to find models that are far better than those found by any other baseline. Furthermore, we find this performance advantage to hold consistently for most other HPO tasks we tried. This empirical demonstration of Mulch's performance is perhaps unsurprising given that all the other methodologies use black-box optimization, while Mulch is able to learn and exploit structure in the optimization problem. 

\subsection{Cost Aware Optimization} 
\label{sec:cost_aware_optimization_es}
We evaluate Mulch-MF on our suite of learning tasks. 
We select two fidelity values of $\rlow$: 0.1 and 0.25 for our algorithm.
We compare against the same baselines as above, but replace Hyperopt with Hyperband \cite{li2017hyperband}, a popular multifidelity HPO algorithm.
All other baselines can only tune XGBoost at the full fidelity.

To make a fair comparison between single-fidelity and multifidelity HPO algorithms, we allow multifidelity methods to use a fractional budget. 
In other words, XGBoost models trained on the full training set use a budget of 1, and models trained on partial data use a budget of $r$, where $0 < r < 1$. Therefore, multifidelity methods can have more than 50 evaluation iterations.

\begin{table*}[!ht]
\small
\centering
\caption{Final accuracy and the total time cost (normalized by the time cost of random search) with a given fixed total budget, averaged over our suite of learning tasks, with lower and upper quartiles. \label{table: final accuracy} }
\begin{tabular}{r| rrrrrrr}
\toprule
\multicolumn{1}{l}{} & \multicolumn{1}{c}{Mulch-MF-0.25} & \multicolumn{1}{c}{Mulch-MF-0.1} & \multicolumn{1}{c}{Hyperband} & \multicolumn{1}{c}{SigOpt} & \multicolumn{1}{c}{Optuna} & \multicolumn{1}{c}{Skopt} & \multicolumn{1}{c}{Random Search} \\ \midrule
Upper quartile (accuracy)      & 94.70\% & \textbf{95.32\%} & 94.72\% & 95.02\%        & 94.80\%        & 94.99\%       & 94.32\%     \\
Lower quartile (accuracy)      & \textbf{83.63\%}   & 82.12\% & 81.16\% & 83.00\%        & 83.06\%        & 83.12\%       & 83.08\%     \\
Average accuracy & \textbf{87.46\%}   & 87.32\% & 87.11\% & 86.67\%        & 86.67\%        & 86.72\%       & 86.36\%    \\
\hline
Upper quartile (time cost)      & 1.19     & \textbf{0.93}    & 1.11& 1.73          & 10.11         & 2.21         & 1       \\
Lower quartile (time cost)      & 0.92 & \textbf{0.76}    & 0.88& 1.29          & 6.74          & 1.76         & 1       \\
Average time cost & 1.01     & \textbf{0.85}    & 0.98& 1.43          & 7.96          & 1.87         & 1      \\
\bottomrule
\end{tabular}
\end{table*}
 
\tabref{table: final accuracy} compares the best accuracy each algorithm reaches averaged over our suite of learning tasks. 
We additionally present the lower and upper quartiles. 
Mulch-MF-0.1  shows the best upper quartile; Mulch-MF-0.25 performs better on average and shows the best lower quartile, i.e., the best ``worst case'' performance. 
The improvement from Mulch-MF-0.1 to Mulch-MF-0.25 is expected since the latter uses more training data in the low fidelity setting.
We also compare how efficient each algorithm is at finding the best hyperparameters.
The bottom of \tabref{table: final accuracy} compares the total time each algorithm takes with the given total budget of 50. 
When comparing time savings, we use the 8 most expensive learning tasks to illustrate the advantages of Mulch-MF.
We normalize the total training cost to that of random search's; a cost of less than 1 means the algorithm is more efficient than random search and vice versa.
Mulch-MF-0.1 is the fastest, and Mulch-MF-0.25 performs comparably to the other multifidelity algorithm Hyperband. All multifidelity algorithms perform better than single-fidelity ones, which shows the effectiveness of multifidelity HPO. 

\subsection{Early Stopping}
Mulch's advantages amplify when we factor in early stopping. For simplicity, we set $n_{s}$ to 10. We re-run Mulch (with metalearning) using early stopping, and find that it saves an average of 85\% compute time over our suite of learning tasks compared to Mulch without early stopping with only a 0.5\% decrease in accuracy. 

Next, we compare Mulch with early stopping and metalearning to other open source software. We quantify our total performance boost as a relative savings over the next best method's total compute time in 50 iterations. For each one of the learning tasks in \tabref{tab:datasets}, we consider the best baseline's mean performance and identify the fraction of time Mulch with early stopping needed to reach that performance. This fraction is often very small. In fact, averaged over our learning tasks, we found that \textit{Mulch is consistently about an order magnitude faster} than model agnostic HPO software. 

Note that the performance gains of early stopping are dependent on the search space for \texttt{num\_boost\_round}. Digging a little deeper, the significant time savings from the early stopping experiments might suggest that we are boosting more than necessary for the HPO tasks, and that the upper bound for \texttt{num\_boost\_round} might be too large. In fact, we chose a maximum of 500, which is reasonable given that the XGBoost default is 100, and is on the low end considering other works that suggest setting the upper bound anywhere from 1000 to 4000 (see \citep{shwartz2021tabular, aws_xgboost_doc, kadra2021tabular}).

\section{Conclusion}
\label{sec:conclusion}
In this paper, we present SigOpt Mulch, a novel hyperparameter tuning system specifically designed for automated tuning of GBTs. Mulch circumvents the standard black-box optimization approach adopted by most hyperparameter software systems today, opting instead for a \textit{model-aware}, \textit{data-driven} optimization process.  

By doing so, Mulch gains two advantages over these traditional HPO systems. First, it is able to automate the process of HPO by making intelligent decisions about the search space, optimization metric, and budget, thus relaxing the domain expertise required to perform HPO. This automation is wrapped in a streamlined but extensible API that is significantly simpler than competitors. Second, Mulch is able to reliably identify better GBT hyperparameters in less time compared to open source HPO software. This is due to our investigations into metalearning and multi-fidelity optimization, which leverage insights about the data, model, and HPO process to significantly boost optimization performance in problems that share the same structure. 

We also shared analysis and algorithmic improvements using metalearning and multifidelity BO to achieve our results.  These analyses are not unique to GBTs and can be generalized to other types of ML models.

\subsection{Limitations and Future Work}
Mulch's primary advantage is simultaneously a disadvantage; what it gains in performance it loses in flexibility. In other words, because it targets specifically GBTs, Mulch cannot perform HPO on other ML models. 
However, given the ubiquity of GBTs, we believe that Mulch possess great value for data scientists and researchers alike. 

There are two other limitations worth mentioning. First, due to the metalearning procedures Mulch uses, we did identify datasets for which Mulch had suboptimal performance. In one such situation, the dataset was larger than anything that Mulch had seen before (around 20 gigs), thus explaining why it didn't perform as well as expected. Problematically, HPO in these regimes is underexplored, due to the computational infeasibility of repeatedly re-running HPO experiments on very large datasets. We believe \textit{data-aware} HPO may pose a solution. In data-aware HPO, we also account for dataset properties such as size or dimension, in the hopes that HPO performance can better extrapolate to datasets of unseen size. 

Second, the scope of this paper is limited to balanced  classification problems models. We do not expect these results to be optimal for highly imbalanced classification problems, especially when classification accuracy is simply not an appropriate measure of model performance. In the future, we plan to repeat the analysis techniques in this article on other such models and problem settings. 


\bibliographystyle{elsarticle-num-names} 
\bibliography{references}

\newpage
\appendix
\section{Mulch Model Tracking}
\label{sec:appendix_mulch_model_tracking}

\begin{figure}[ht!]
    
    
\caption{Mulch Run API (\texttt{sigopt.xgboost.run}) for model tracking.}
\label{fig:xgb_run_api}
\end{figure}

\begin{figure}[ht!]
    \centering
    \includegraphics[width=0.45\textwidth]{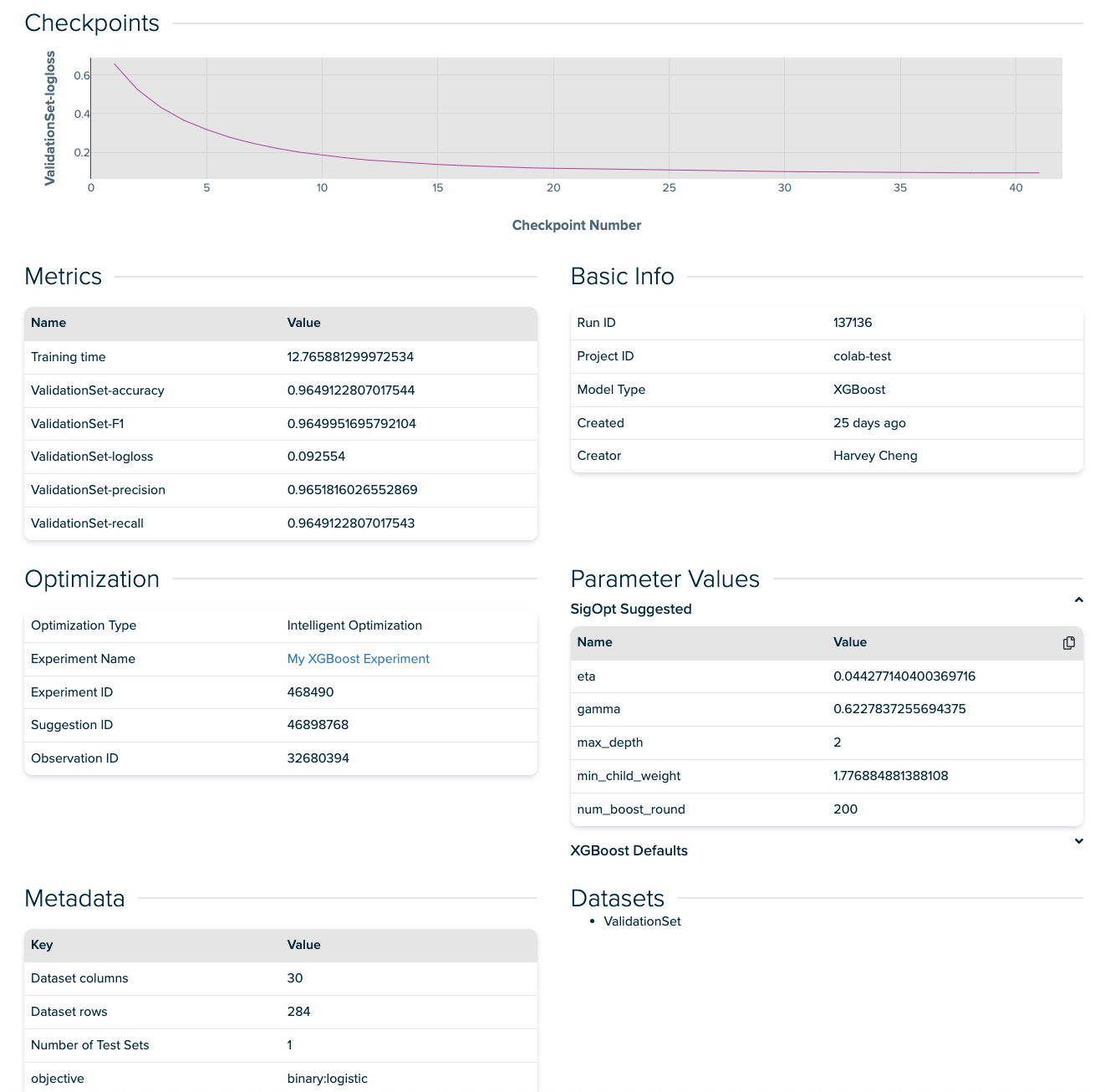}
    \caption{Screenshot of the analysis page for one of the XGBoost models within a SigOpt Mulch experiment.}
    \label{fig:run_page}
\end{figure}

\noindent
The building blocks of Mulch are the Run API and Experiment API. 
The Mulch Run API wraps around the XGBoost training API (\texttt{xgboost.train}) and the vanilla SigOpt \texttt{run()} functionality.
As shown in \figref{fig:xgb_run_api}, the Run API mirrors the XGBoost training API by design.
Functionally, it trains the XGBoost model and automatically logs all relevant information, such as hyperparameter values, evaluation metrics, metadata, and learning curves, to SigOpt without additional code. 
\figref{fig:run_page} shows a screenshot of this augmented model tracking. 
Since the Mulch Experiment API is built on top of the Run API, every model tuned with Mulch enjoys the automatic tracking capabilities.

\subsection{Additional tests}

\begin{table*}[t]
    \centering
            \caption{We test Mulch on large datasets, representing real-world scenarios that might be encountered by data scientists and machine learning practitioners. Mulch almost always achieves superior prediction accuracy in less time vanilla SigOpt. To be precise, we estimated Mulch to \textit{31 percent faster} than the default SigOpt optimization engine with an \textit{8} percent improvement in performance. 
}\label{tab:production_testing}
    \begin{tabular}{r  c c | c c c c c }
    \toprule
        \textbf{Dataset} & \textbf{Size} & \textbf{Mulch} & \textbf{Duration} & Improve & \textbf{HPO Accuracy}  & Improve  \\ \midrule
        \multirow{2}{*}{\texttt{guillermo}} & \multirow{2}{*}{600 MB }  & $\highlightthree{\checkmark}$ & \textbf{33 min} & \multirow{2}{*}{\highlightthree{$+$48 \%}} & 0.8394 & \multirow{2}{*}{\highlightfour{$-4 \%$}} \\ 
                               && \highlightfour{\text{\sffamily X}}  & 64 min & &\textbf{0.8404}  & \\ \hline
        \multirow{2}{*}{\texttt{hill-valley}}  & \multirow{2}{*}{30 MB}    & $\highlightthree{\checkmark}$ & \textbf{33 sec} & \multirow{2}{*}{\highlightthree{$+$82 \%}}& \textbf{0.6204 }   & \multirow{2}{*}{\highlightthree{$+6 \%$}} \\ 
                               && \highlightfour{\text{\sffamily X}}  & 3 min  &  &0.6171     & \\ \hline
        \multirow{2}{*}{\texttt{real-sim}}     & \multirow{2}{*}{90 MB}     & $\highlightthree{\checkmark}$ & \textbf{17 min} & \multirow{2}{*}{\highlightthree{$+$15 \%}}& \textbf{0.9673}     & \multirow{2}{*}{\highlightthree{$+ 23 \%$}}\\ 
                               && \highlightfour{\text{\sffamily X}}  & 20 min & & 0.9663     & \\ \hline
        \multirow{2}{*}{\texttt{rcv1}}         & \multirow{2}{*}{1.3 GB}    & $\highlightthree{\checkmark}$ & \textbf{3.8 hours}  & \multirow{2}{*}{\highlightthree{$+$49 \%}}& \textbf{0.9836}     & \multirow{2}{*}{\highlightthree{$+2 \%$}} \\ 
                               && \highlightfour{\text{\sffamily X}}  & 7.5 hours  &  & 0.9831    & \\ \hline
        \multirow{2}{*}{\texttt{usps}}         & \multirow{2}{*}{33 MB}    & $\highlightthree{\checkmark}$ & 14 min & \multirow{2}{*}{\highlightfour{ $-$17 \%}}& \textbf{0.9673}     & \multirow{2}{*}{\highlightthree{$+3 \%$}}\\ 
                               && \highlightfour{\text{\sffamily X}}  & \textbf{12 min} & & 0.9668     & \\ \hline
        \multirow{2}{*}{\texttt{yahoo-ltrc}}   & \multirow{2}{*}{1.7 GB}   & $\highlightthree{\checkmark}$ & \textbf{7.7 hours} & \multirow{2}{*}{\highlightthree{$+$11 \%}} & \textbf{0.6857}  & \multirow{2}{*}{\highlightthree{$+15 \%$}} \\ 
                               && \highlightfour{\text{\sffamily X}}  & 8.6 hours  & & 0.6849  & \\ 
                               \bottomrule

    \end{tabular}
\end{table*}

In this section, we provide some experiments involving production-level settings: large datasets and long training times. In this setting, we do not perform hyperparameter optimization across all optimization packages and for multiple replications, given the prohibitive nature of such experiments; we simply run optimization using SigOpt with and without Mulch, and observe the performance difference and time savings. 

Training is performed using an AWS \texttt{c5.24xlarge} instance with 96 CPUs and 200 GB of memory, representing a real-world production system that might be used by data scientists and machine learning practitioners to train and deploy large gradient boosted trees. We considered datasets both small and large for this test including the popular Yahoo learning to rank (\texttt{yahoo-ltrc}) dataset from the original XGBoost paper \cite{chen2016xgboost}. 

In \tabref{tab:production_testing}, we test on six datasets both with and without Mulch. For each HPO experiment, we provide the base accuracy (the accuracy at the start of optimization) and the end performance for HPO. We also list the duration of experiments, and the improvement in both wall clock time and HPO accuracy. We compute these with the following formulas, respectively:

\[
\text{Time Improvement} = 1 - \frac{\text{Duration}(Mulch)}{\text{Duration}(No \; Mulch)},
\]

\begin{align*}
&\text{Accuracy Improvement} = \\ 
&\frac{\text{Accuracy}(Mulch) - \text{Accuracy}(Base)}{\text{Accuracy}(No \; Mulch)- \text{Accuracy}(Base)}.
\end{align*}

Mulch almost always achieves superior prediction accuracy in less time than vanilla SigOpt. On average, it is \textit{31 percent faster} and demonstrates an \textit{8} percent improvement in performance.

\end{document}